\documentclass[10pt,twocolumn,letterpaper]{article}

\usepackage{cvpr}              

\usepackage{times}
\usepackage{epsfig}
\usepackage{graphicx}
\usepackage{amsmath}
\usepackage{amssymb}
\usepackage{booktabs}
\usepackage{float}
\usepackage{color,xcolor}
\usepackage{colortbl}
\usepackage{multirow}
\usepackage{bbding}

\usepackage[accsupp]{axessibility}

\usepackage[pagebackref,breaklinks,colorlinks]{hyperref}

\usepackage[capitalize]{cleveref}
\crefname{section}{Sec.}{Secs.}
\Crefname{section}{Section}{Sections}
\Crefname{table}{Table}{Tables}
\crefname{table}{Tab.}{Tabs.}

\def\cvprPaperID{2364} 
\def\confName{CVPR}
\def\confYear{2022}

\begin{document}

\title{Location-Free Human Pose Estimation}
\author{
Xixia Xu$^{1}$\footnotemark[1],\quad Yingguo Gao$^{2}$,\quad Ke Yan$^{2}$\footnotemark[2],\quad Xue Lin$^{1}$,\quad Qi Zou$^{1}$\footnotemark[2]\\
$^{1}$Beijing Key Laboratory of Traffic Data Analysis and Mining, Beijing Jiaotong University, China\\
$^{2}$Tencent Youtu Lab, Shanghai, China\\
{\tt\small \{19112036,18112028,qzou\}@bjtu.edu.cn; \quad \{yingguogao,kerwinyan\}@tencent.com}
}

\maketitle
\renewcommand{\thefootnote}{\fnsymbol{footnote}}
\footnotetext[1]{Works done while interning at Tencent Youtu Lab.}
\footnotetext[2]{Corresponding author.}

\begin{abstract}
  Human pose estimation (HPE) usually requires large-scale training data to reach high performance. However, it is rather time-consuming to collect high-quality and fine-grained annotations for human body. To alleviate this issue, we revisit HPE and propose a location-free framework without supervision of keypoint locations. We reformulate the regression-based HPE from the perspective of classification. Inspired by the CAM-based weakly-supervised object localization, we observe that the coarse keypoint locations can be acquired through the part-aware CAMs but unsatisfactory due to the gap between the fine-grained HPE and the object-level localization. To this end, we propose a customized transformer framework to mine the fine-grained representation of human context, equipped with the structural relation to capture subtle differences among keypoints. Concretely, we design a Multi-scale Spatial-guided Context Encoder to fully capture the global human context while focusing on the part-aware regions and a Relation-encoded Pose Prototype Generation module to encode the structural relations.
  All these works together for strengthening the weak supervision from image-level category labels on locations. Our model achieves competitive performance on three datasets when only supervised at a category-level and importantly, it can achieve comparable results with fully-supervised methods with only $25\%$ location labels on MS-COCO and MPII. 

\end{abstract}

\section{Introduction}
\label{sec:intro}
Human pose estimation (a.k.a., keypoint localization) is a challenging yet fundamental computer vision task, which aims to detect the keypoint locations (\emph{e.g.}, eyes, ankles). In recent years, HPE has witnessed dramatic progress with the development of CNNs. An integral factor of the achievement is the availability of large-scale training data with precise location annotations. However, it's rather label-intensive and time-consuming to collect high-quality and fine-grained annotations. Thus, we study the keypoint localization when only the image-level category labels are given.

\begin{figure}[t]
	\centering
	\setlength{\abovecaptionskip}{0.23cm}
	\setlength{\belowcaptionskip}{0cm}
	\includegraphics[width=3.1in,height=1.9in]{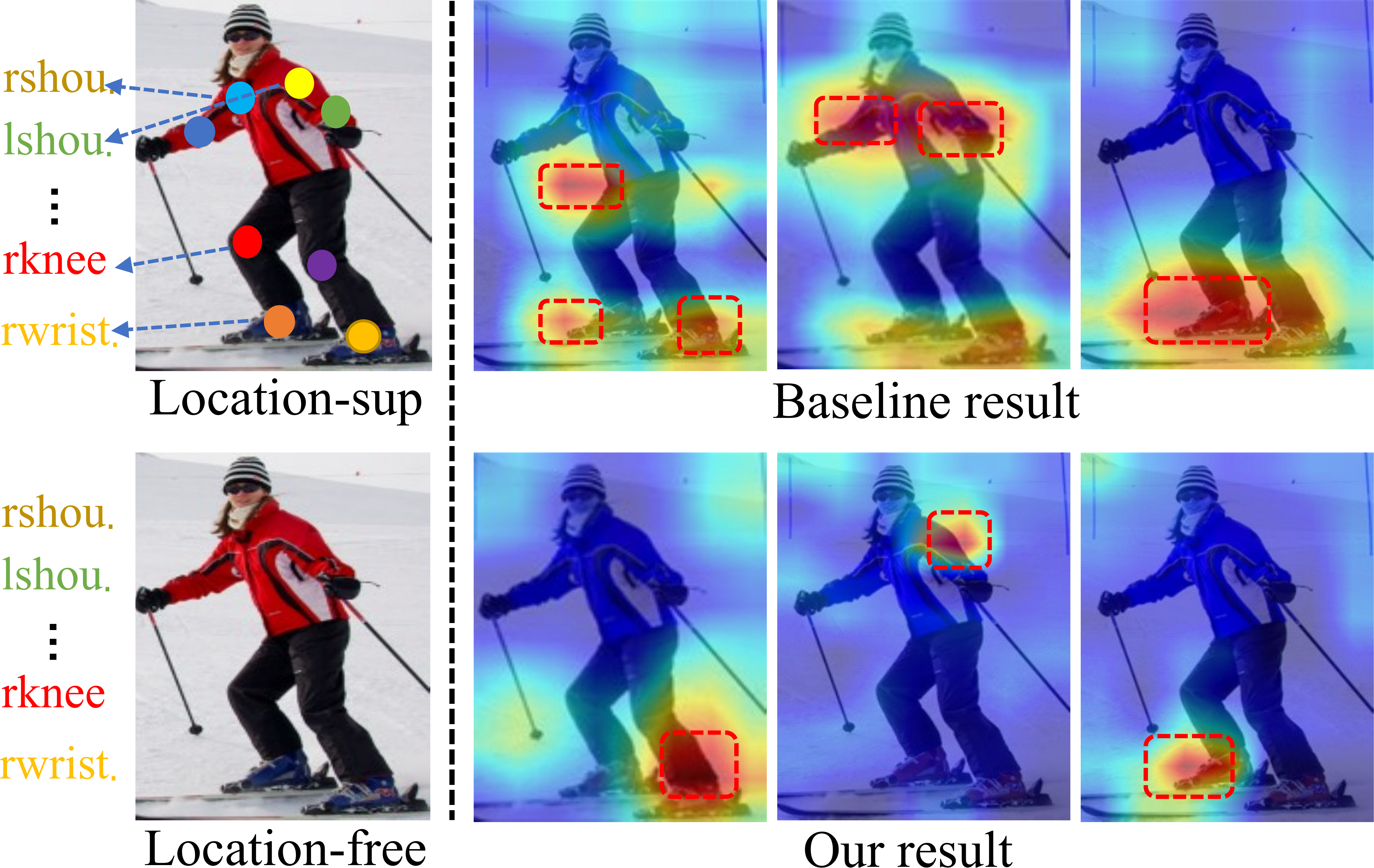}
	\caption{Column 1: \textbf{Location-supervised v.s. Location-free} HPE; Column $2$-$4$: the result of location-free baseline v.s. the result of our method. Noted that the correct joints are lankle, lshou and rankle respectively.}
	\label{fig:intro}
\end{figure}

The Class Activation Mapping (CAM)\cite{zhou2016learning} is a simple but effective method to discover object regions from intermediate classifier activation with only image-level labels, which has been the cornerstone of weakly-supervised object localization (WSOL)\cite{zhou2016learning} and weakly-supervised semantic segmentation (WSSS)\cite{zhang2021complementary}. The CAM tends to focus on the most discriminative parts of object, and many approaches\cite{kumar2017hide, yun2019cutmix, zhang2020inter} are proposed to improve CAM to cover the full extend of an object. But these methods cannot localize the subtle joints due to the gap between the object-level localization and fine-grained keypoint localization.

Our method is also built upon CAM. There are two main obstacles for applying CAM in HPE. \textbf{i)} For tiny and local keypoints, it's hard for the model to capture the precise spatial features for accurate prediction without the explicit location labels. Furthermore, the local appearance features learned only relying on the image-level labels are not comprehensive enough for understanding the human body. Therefore, more contextual information and explicit spatial prior need to be exploited. \textbf{ii)} The inter-class differences among joints are subtle and the adjacent or symmetric joints possess similar semantic context or appearances. It tends to result in the location confusions and incorrect responses as in Fig.~\ref{fig:intro} (Baseline result in column $2$-$4$). It's hard for model to capture the fine-grained joint-specific features to eliminate the confusions especially without the explicit location supervision. The inherent structural relation between joints plays a critical role in helping distinguish or infer the uncertain locations. Thus, how to dig out the intrinsic structural relation prior for the model is vital.

\begin{figure*}[htb]
	\centering
    \includegraphics[width=6.1in,height=2.65in]{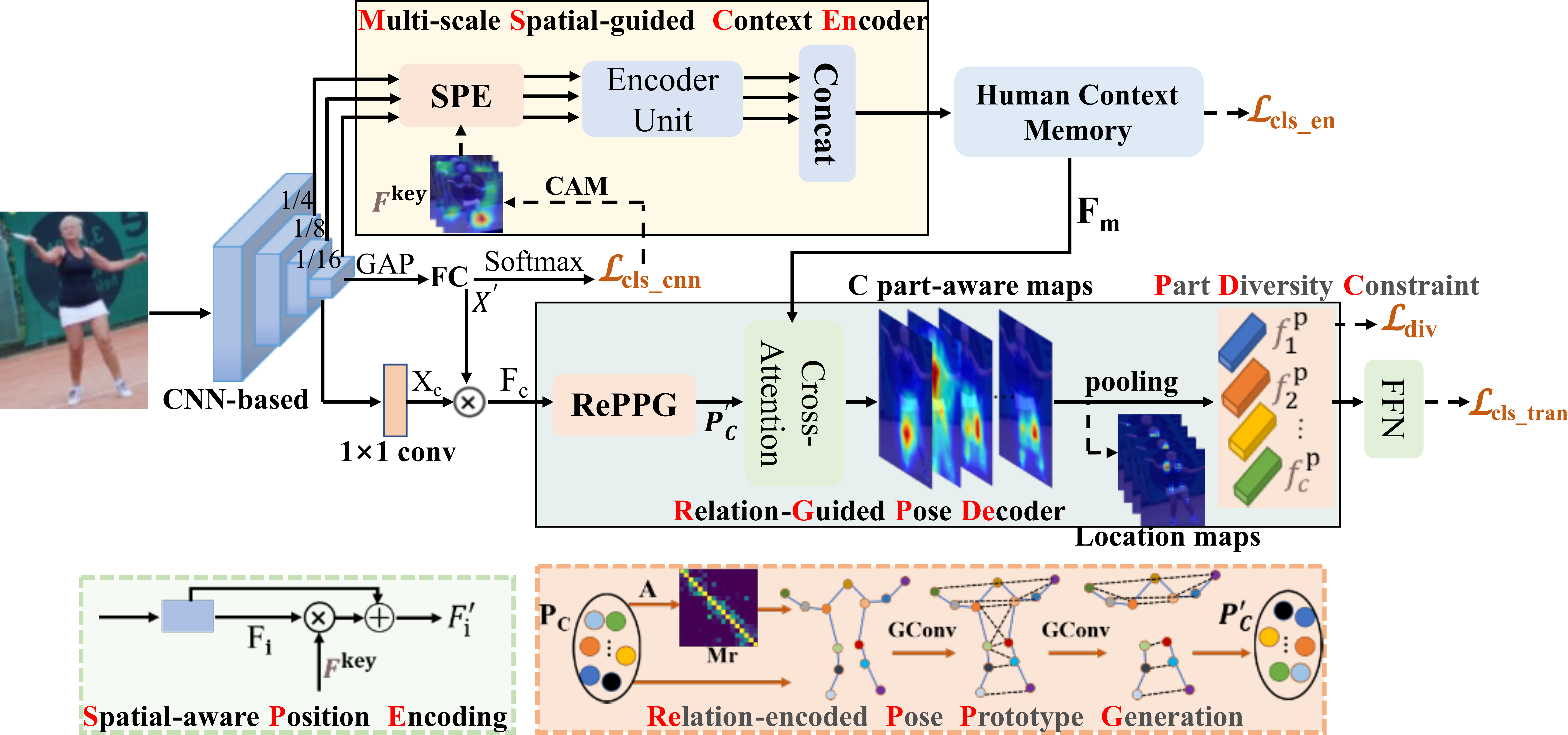}
	\caption{\textbf{LO}cation-\textbf{FR}ee \textbf{(LOFR)} HPE pipeline. It comprises a \textbf{MSC-En} with \textbf{SPE} to capture the multi-scale human context related to the part regions and a \textbf{RGP-De} integrated with \textbf{RePPG} that generate the encoded pose prototypes guiding the cross-attention with human context to parse more accurate part responses. The \textbf{PDC} helps learn more distinguished part features.}
	\label{fig:2network}
\end{figure*}

Based on above discussions, we propose a novel customized transformer-based architecture for \textbf{LO}cation-\textbf{FR}ee \textbf{(LOFR)} HPE as in Fig.~\ref{fig:2network}. Firstly, thanks to the self-attention mechanism in Transformer\cite{vaswani2017attention}, the global contextual information can be effectively captured in HPE. To better capture the precise spatial information, we propose a \textbf{M}ulti-scale \textbf{S}patial-guided \textbf{C}ontext \textbf{En}coder \textbf{(MSC-En)}. In MSC-En, we design a \textbf{S}patial-aware \textbf{P}osition \textbf{E}ncoding \textbf{(SPE)} module helps the model focus on body regions while capturing the global human context. We capture the multi-scale feature representations to conduct self-attention learning leading to more comprehensive context from aggregated multi-scale information, which is robust to background clutters. For alleviating the location confusions, we equip the model with the structural relations encoded via GCN and propose a \textbf{R}elation-\textbf{G}uided \textbf{P}ose \textbf{De}coder \textbf{(RGP-De)}. In RGP-De, a \textbf{Re}lation-encoded \textbf{P}ose \textbf{P}rototype \textbf{G}eneration \textbf{(RePPG)} module is designed to express keypoint-specific relations to help infer the confused parts during decoding. Finally, part-aware response maps denoting the spatial distributions of specific keypoints (\emph{e.g.}, ankle or head) are acquired by exploring the interactions between human context memory and prototypes. To prompt diversity and fine-grain, a \textbf{P}art \textbf{D}iversity \textbf{C}onstraint \textbf{(PDC)} is devised to encourage lower correlation between part features and force them focus on their own parts.

In a nutshell, the contribution of this paper is three-fold:

\begin{itemize}
   \item To our knowledge, we are the first to develop a location-free HPE only with image-level labels. The effectiveness is extensively validated on three datasets and the performance can even outperform the supervised one when few location labels are given.
   \item We employ a multi-scale spatial-guided context encoder to capture the global context features and meanwhile make it attend to the body parts by the aid of the spatial-aware positional encoding. 
   \item We design a relation-encoded pose prototype generation strategy to mine the inherent spatial relation prior between keypoints via GCN. Also, a part diversity constraint makes the part-aware features more distinguished. 
\end{itemize}

\section{Related Work}
\noindent\textbf{Human Pose Estimation.}
Recently, researchers have made painstaking efforts\cite{yang2017learning,sun2017human,chu2017multi-context} to make progress on HPE, two mainstream methods are prevalent including bottom-up\cite{papandreou2017towards,kocabas2018multiposenet:} and top-down \cite{chen2018cascaded,xiao2018simple,sun2019deep}. The former directly estimate all the keypoints and assign them into persons. The latter firstly detect the human bounding boxes and locate the keypoints within each box. However, the above works all solve the fully-supervised regression and no research explore the weakly-supervised HPE. This paper follows the top-down pipeline. After acquiring the bounding boxes, we acquire the keypoint locations from a perspective of classification with only the category labels.

\noindent\textbf{Transformers in computer vision.}
Recently, Transformers attract much attention in computer vision. ViT\cite{dosovitskiy2020image} applies a pure Transformer framework to a series of image patches for classification. Besides, vision Transformer is widely applied to object detection\cite{zheng2020end}, segmentation\cite{zheng2021rethinking}. Further, DETR\cite{carion2020end} and Deformable DETR\cite{zhu2020deformable} predict a box set for matching the object location.

Specially, the Transformer is also applied in HPE, including hand pose estimation\cite{huang2020hot} and 3D HPE\cite{li2021lifting,zheng20213d}. The most close to ours is the applications in 2D HPE\cite{li2021pose,mao2021tfpose,yang2021transpose,li2021tokenpose}. These studies achieve impressive performance implying that Transformer is suitable and effective for modeling human poses. Therefore, we also leverage the transformer to explore the weakly-supervised HPE with only the category labels.

\noindent\textbf{CAM-based WSOL.} Weakly-supervised object localization aims to locate the objects with only image-level labels. Since CAM was proposed in \cite{zhou2016learning}, CAM-based methods have been achieved great success for both WSOL and WSSS. For WSOL, CAM suffers from only identifying small discriminative region of objects. After that, a series of works \cite{kumar2017hide,yun2019cutmix,zhang2020inter} are proposed to improve the quality of CAM. However, these improvements are ineffective when extended to HPE.

In light of CAM, we rethink the HPE and aim to locate keypoints with the image-level category labels. Even if CAM-based methods have made success in WSOL, it doesn't work well for the keypoint prediction. The local human parts and the subtle differences between categories undoubtedly bring great challenges to achieve accurate HPE.

\section{Method}
\label{method}
\subsection{Framework Overview} As shown in Fig.~\ref{fig:2network}, the LOFR framework primarily comprises of the MSC-En and RGP-De. Given an input, we firstly obtain the multi-scale feature representations through the CNN backbone. Then, the multi-level feature maps are processed with SPE to feed into the encoder to conduct the self-attention for capturing the human context memory $F_{m}$ around body regions. In RGP-De, a set of pose prototypes $P^{'}_{C}$ that initialized with RePPG are sent into decoder performing cross-attention with the context memory decoding the part-aware response maps. Part features can be obtained by pooling operation where the location maps are treated as different spatial keypoint locations. Except the general binary cross entropy (BCE) loss for the predicted joint categories, a part diversity constraint is used to capture more distinguished part features.

\subsection{CNN-based}
\label{cnn}
Following the top-down pipeline, the detected single person images are fed into the CNN-based network to acquire the feature map $X\in\mathbb{R}^{ H\times W\times D}$ where $H, W$, and $D$ are the height, width and channels. We turn $X$ into $X'\in\mathbb{R}^{ H\times W\times C}$ with $1\times1$ convolution and $C$ is the number of keypoint classes. Then, followed by  GAP, FC and a softmax layer for classification as in Fig.~\ref{fig:2network}, we get the category-specific activation maps $M = [m_{1},m_{2},...,m_{c}]\in \mathbb{R}^{ H\times W\times C}$ by convolving the weights of FC with feature map $X'$ as follows:
\begin{equation}
\begin{aligned}
\label{cam}
m_{c}(x,y) = RELU(\sum_{k}w^{c}_{k}x'_{k}(x,y)),
\end{aligned}
\end{equation}
where $w^{c}$ and $x'_{k}$ depict the weight of the $c$-th category and $k$-th feature map, respectively.

For acquiring the initial node embedding for the graph in RGP-De (in sec \ref{RGP-De}), we compute category-specific keypoint vectors $F_{c} = [f_{1},f_{2},...,f_{c}] \in \mathbb{R}^{ C\times D'}$ as: 
\begin{equation}
\begin{aligned}
\label{node feature}
f_{c} = x'^{T}X^{c} = \sum_{i=1}^{H}\sum_{j=1}^{W}x'_{i,j}x^{c}_{i,j},
\end{aligned}
\end{equation}
where $x'_{i,j}$, $x^{c}_{i,j}$ are the $(i,j)$-th feature of feature maps and $X^{c} \in\mathbb{R}^{ H \times W\times D'}$ is got with a $1\times1$ convolution.

\subsection{Multi-scale Spatial-guided Context Encoder}
We propose a MSC-En equipped with the SPE to capture the multi-scale spatial-aware human context information as shown in the yellow box of Fig.~\ref{fig:2network}.

\noindent\textbf{Spatial-aware Position Encoding.}
As discussed above, we have obtained the part-aware CAMs $M$. Further, we obtain coarse keypoint location maps $\{F^{key}_{i}\}, i\in{1,2,...C}$ via finding the max value of $M$. Rather than adopt the random initialized position embedding, we take $\{F^{key}\}$ as the implicit position prior to help the model know spatial part locations while capturing the context. Given the input features $F$, we establish spatial-aware inputs as below:
\begin{equation}
\begin{aligned}
\label{PE}
\varphi = F \otimes F^{key},
\end{aligned}
\end{equation}
\begin{equation}
\begin{aligned}
\label{PE1}
F' = \psi(\varphi \oplus F).
\end{aligned}
\end{equation}
We view $\varphi$ as the updated positional encoding to sum with $F$, the $\psi$ depicts a feature transformation operation. The $\otimes, \oplus$ depicts the cross-product and element-wise sum operation. The resulted $F'$ is fed into the encoder.

\noindent\textbf{Multi-scale Context Learning.} For capturing more comprehensive context, we extract the multi-scale features $\{F_{i}\}\in \mathbb{R}^{ H\times W\times C}, i\in{1,2,3}$ from the CNN backbone at the down-sampled ratios of $1/4, 1/8, 1/16$. We obtain their position encoding and the updated multi-scale input features $\{F'_{i}\}, i\in{1,2,3}$ in Eq.~\ref{PE} to conduct self-attention (SA) mechanism. 
\begin{figure}[t]
	\centering
	\includegraphics[width=3.36in,height=0.75in]{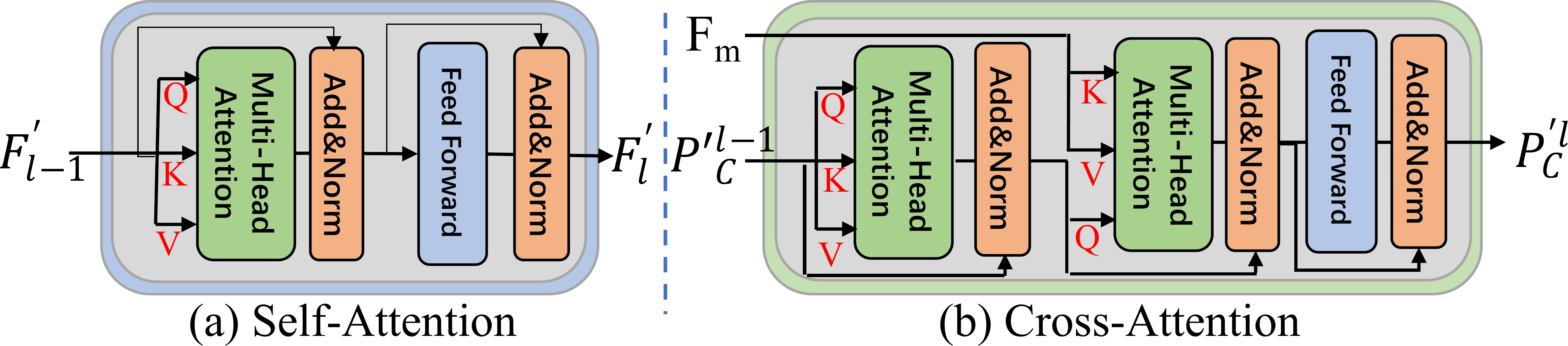}
	\caption{Illustration of the Encoder-Self-Attention and Decoder-Cross-Attention modules.}
	\label{fig:module}
\end{figure}

Given $F'$ as input, the multiple SA modules are employed to learn pixel-level human contextual dependencies of multi-scale features. As depicted in Fig.~\ref{fig:module} (a), a SA module consists of multi-head self-attention (MSA) and feed forward networks (FFN), layer normalization (LN) and residual connection is applied after every block. The FFN contains two linear layers with a ReLU. For $l$-th layer, the input to SA is a triplet of (query, key, value) computed from the input $F'_{l-1}$ as:
\begin{equation}
\begin{aligned}
\label{qkv}
Q = F'_{l-1}W^{l}_{Q}, K = F'_{l-1}W^{l}_{K}, 
V = F'_{l-1}W^{1}_{V},
\end{aligned}
\end{equation}
where $W^{l}_{Q}\in \mathbb{R}^{ C\times d_{q}}$, $W^{l}_{K}\in \mathbb{R}^{ C\times d_{k}}$, $W^{l}_{V}\in \mathbb{R}^{ C\times d_{v}}$ are the parameter matrices of linear projection heads, and $d_{q}$, $d_{k}$ and $d_{v}$ is the dimensions of input. The SA is formulated as:
\begin{equation}
\begin{aligned}
\label{weight}
s_{qk} = Softmax(\frac{F'_{l-1}W^{l}_{Q}(F'_{l-1}W^{l}_{K})^{T}}{\sqrt{d_{k}}}),
\end{aligned}
\end{equation}
where the attention weights $s_{qk}$ are calculated based on the dot-product similarity between each query and the keys. The $d_{k}$ is a scaling factor modeling the inter-dependency between different spatial pixels of human part regions.
\begin{equation}
\begin{aligned}
\label{SA}
SA(F'_{l-1}) = F'_{l-1} + F'_{l-1}W^{l}_{V}s_{qk},
\end{aligned}
\end{equation}
the weighted sum of the values can aggregate these semantically related spatial pixels to update the context. Since pixels belonging to the human body part have high similarities while are distinct from background pixels, the feature captures more complete human body would be more robust to backgrounds. The MSA is an extension with $h$ SA and projects their concatenated output as:
\begin{equation}
\begin{aligned}
\label{MSA}
MSA(F'_{l-1}) = Concat(SA_{1},...,SA_{h})W^{l}_{O},
\end{aligned}
\end{equation}
where $W^{l}_{O}\in \mathbb{R}^{hd_{k} \times C}$ is a parameter of linear head. We set $h = 8, C = 256$ and $d_{q}$,$d_{k}$,$d_{v}$ equal to $C/h = 32$. Then, we use FFN to produce the context-aware memory:
\begin{equation}
\begin{aligned}
\label{output}
F'_{l} = MSA(F'_{l-1}) + FFN(MSA(F'_{l-1}))\in \mathbb{R}^{L\times C}.
\end{aligned}
\end{equation}
Up to now, the single-scale human context features are captured, which are robust to background clutters. For aggregating the multi-scale context, we concat the multi-level output $\{F'_{i}\}$, $i\in\{1,2,3\}$ as the final context memory $F_{m}$.

\subsection{Relation-Guided Pose Decoder}
\label{RGP-De}
In RGP-De, we involve a set of relation-encoded pose prototypes that contain the category-specific semantics and structural relations among joints to let the decoder decode more accurate part-aware response maps. An additional part diversity constraint enables the part locations more accurate and focused.

\noindent\textbf{Relation-encoded Pose Prototype Generation.} Notably, the human poses have inherent structure conforming to the kinematic constraint. For example, the adjacent or symmetrical joints are more likely to possess highly consistent semantic information. Therefore, we design RePPG to integrate joint-wise relations into the updated pose prototypes for parsing more precise part-aware locations.

We firstly introduce a set of pose prototypes $\mathcal{P}_{C} = \{{p}_{i}\}^{C}_{i=1}, p_{i}\in \mathbb{R}^{ 1\times d}$ which determines whether pixels of the feature $F_{m}$ belong to the part $i$. We initialize the category-specific keypoint features $F_{c}$ in Eq.~\ref{node feature} as the node feature of $\mathcal{P}_{C}$. We build an intuitive graph $G = (V,E)$ based on $F_{c}$. $V = \{{v_{i}, i = 1, 2, ..., C}\}$ is the node set to depict keypoints. $E =$ $\{$${v_{i}v_{j}}$ $|$ if $i$ and $j$ are connected in the human body$\}$ refers to limbs of part. The adjacent matrix $A =$ $\{$$a_{ij}$$\}$ is initialized according to the pre-defined kinematic connection with $a_{ij} = 1$ when $v_{i}$ and $v_{j}$ are adjacent in $G$ or $i = j$, otherwise $a_{ij} = 0$. 

Considering the human body structure is a natural graph with spatial constraint among joints, we thus model the keypoint relations via the recent SemGCN\cite{zhao2019semantic} to explore their structural relation. For a graph convolution, propagating features through neighbors helps to learn robust structure and captures the joint dependency. This equips pose prototypes with the structural relation prior, which is vital for activating the uncertain part responses while decoding.

Concretely, the updated node features are firstly gathered to node $i$ from its neighbors $j$. The initial node features are collected into ${F_{c}}^{(l)} \in R^{D_{l}\times C}$ as in Eq.~\ref{node feature}. 
\begin{equation}
\begin{aligned}
\label{graph}
{F_{c}}^{(l+1)} = \vec{W}^{(l+1)}{F_{c}}^{(l)}\varphi_{i}(M_{r}\odot A),
\end{aligned}
\end{equation}
where ${F_{c}}^{(l)}$ and ${F_{c}}^{(l+1)}$ are the node features before and after $l$-th conv., $\varphi_{i}$ is normalization, $\vec{W}^{(l+1)}$ is the weight matrix. The $M_{r} \in R^{C \times C}$ denotes the local semantic relations between joints and updates with node features. In this way, the updated pose prototypes $\mathcal{P}^{'}_{C} = \{{p}^{'}_{i}\}^{C}_{i=1}$ encoding the local semantic and spatial relation are obtained.

The cross-attention layer aims to learn more specific part-aware features through the interaction between $F_{m}$ with the prototypes $\mathcal{P}^{'}_{C}$. As shown in Fig.~\ref{fig:module} (b), given $F_{m}$, queries come from prototypes $\{{p}^{'}_{i}\}^{C}_{i=1}$, keys and values arise from the feature $F_{m}$. The implementation is the same as the above SA learning, the attention weights of all $hw$ positions form a part-aware response map $R_{i} = [r_{i,1},r_{i,2},r_{i,3},...,r_{i,hw}]$, which has high response values at context features belonging to $i$-th part. We then obtain $i$-th part feature by pooling operation. By computing over all prototypes, we obtain $C$ part response maps (each map is an attention map) and the corresponding part features $\{F^{p}_{i}\}^{C}_{i}$. Finally, we obtain the keypoint location map via finding the max value of their response map.

\noindent\textbf{Part Diversity Constraint.} A simple classification loss is insufficient to capture subtle differences among keypoint categories. Different prototypes may tend to focus on the same part (\emph{e.g.}, the main body), which may result in confusion among keypoint locations. Thus, making part features more distinguished, we design a part diversity constraint on $\{F^{p}_{i}\}^{C}_{i}$ to let features attend to the corresponding local part. 
\begin{equation}
\begin{aligned}
\label{div_loss}
\mathcal{L}_{div} = \frac{1}{C(C-1)}\sum^{C}_{i=1}\sum^{C}_{j=1,i\neq j}\frac{\langle f^{p}_{i}, f^{p}_{j}\rangle}{\lVert f^{p}_{i}\rVert_{2}\cdot\lVert f^{p}_{j}\rVert_{2}},
\end{aligned}
\end{equation}
if the $i$-th and $j$-th part feature give a high weight to the same location, the $\mathcal{L}_{div}$ will become large and prompt each part feature to adjust themselves adaptively.

\subsection{Optimization} The classification output denotes joint-wise one-hots $\mathcal{O}$. Based on this, the overall objective function $\mathcal{L}_{weak}$ includes three classification losses and a diversity loss. The classification loss $\mathcal{L}_{cls}$ refers to the BCE, followed by CNN output, the encoder output and the final prediction, respectively. 
\begin{equation}
\begin{aligned}
\label{bce}
\mathcal{L}_{cls} =  \frac{1}{C}\sum^{C}_{i=c}BCE(\mathcal{O}_{i},\mathcal{O^{*}}_{i}).
\end{aligned}
\end{equation}
\begin{equation}
\begin{aligned}
\label{loss}
\mathcal{L}_{weak} = \!\alpha\mathcal{L}_{cls_{cnn}}\!+\!\alpha_{1}\mathcal{L}_{cls_{en}}\!+\!\alpha_{2}\mathcal{L}_{cls_{tran}}\!+\! \beta\mathcal{L}_{div}.
\end{aligned}
\end{equation}
When we provide few location-labeled data, we adopt $\mathcal{L}^{sup}_{mse}$ to measure the groundtruth $H^{*}$ and the predicted heatmap $H$. The overall loss $\mathcal{L}_{semi_{-}weak}$ is depicted as,
\begin{equation}
\begin{aligned}
\label{sup}
\mathcal{L}^{sup}_{mse} = \frac{1}{C}\sum MSE(H^{*},H),
\end{aligned}
\end{equation}
\begin{equation}
\begin{aligned}
\label{loss_sup}
\mathcal{L}_{semi_{-}weak} = \mathcal{L}^{sup}_{mse} + \mathcal{L}_{weak},
\end{aligned}
\end{equation}
where $\alpha$, $\alpha_{1}$, $\alpha_{2}$ and $\beta$ are the weight factors, respectively. 

\section{Experiment}
\subsection{Datasets and Evaluation metric}
\noindent\textbf{COCO Keypoint Detection}\cite{lin2014microsoft} consists of $118K$ training images, $20K$ testing images and the $5K$ validation images. The performance is evaluated by the OKS-based average precision (AP) and average recall (AR). 

\noindent\textbf{MPII Human Pose Dataset} consists of $25K$ images with $40K$ objects, where $12K$ objects are for testing and the remaining for training. We use the standard PCKh\cite{andriluka20142d} (head-normalized probability of correct keypoint) for evaluation. 

\noindent\textbf{CrowdPose} contains $20$$K$ images and $80$$K$ human instances in three crowding levels by Crowd Index: easy ($0$ $\sim$ $0.1$), medium ($0.1$ $\sim$ $0.8$) and hard ($0.8$ $\sim$ $1$). It aims to promote performance in crowded cases and adopts the same evaluation metrics as in MS-COCO.

\subsection{Implementation Details}
\noindent\textbf{Network Architectures.} Unless specified, the backbone adopts ResNet-$50$ and HR-w$32$. We choose DETR\cite{carion2020end} as the Transformer baseline.

\noindent\textbf{Training.} We implement all experiments in PyTorch\cite{Paszke2017AutomaticDI} on $4$ Tesla V100s with $32$GB. For MS-COCO, human detection boxes are resized to $256$$\times$$192$ or $384$$\times$$288$. We adopt Adam\cite{kingma2015adam:} optimizer with a learning rate of $4\times10^{-3}$ and a weight decay of $10^{-4}$. The learning rate of transformer is decreased by a factor of $10$. For MPII, the input size adopts $256$$\times$$256$ and $384$$\times$$384$ and half-body augmentations are adopted. The training lasts for $160$ epochs. For CrowdPose, the setting is similar with COCO and trained for $210$ epochs. For data augmentation, we apply random flip and random resize with scale $\in$ $[0.65,1.35]$ (cutout not used). 
The weight factors for the $\alpha$, $\alpha_{1}$, $\alpha_{2}$ and $\beta$ are set as $0.2$, $0.2$, $0.5$ and $0.1$, respectively.

\subsection{Comparison with State-of-the-arts}

\subsubsection{Location-free Setting}
\textbf{On MS-COCO.} The result comparisons on MS-COCO test-dev are in Tab.~\ref{COCO val}. It's noted that the baseline is implemented based on the CNN (Res-$50$) with the original transformer-based architecture. The accuracy has sharply dropped compared with the supervised result. By contrast, ours achieves over $20\%$ improvement than the baseline across all cases. This proves our LOFR can obtain more accurate keypoint locations.
Although a certain gap exists with the supervised one, we still achieve a competitive performance especially compared with the bottom-up methods. We also achieve stable improvements even with different backbones and input sizes, which also reflect the good generality of our method. 

\begin{table}[ht]
	\centering
	\setlength{\tabcolsep}{0.48mm}
	\setlength{\abovecaptionskip}{0cm}
	\setlength{\belowcaptionskip}{0cm}
			\caption{Performance comparisons on COCO \emph{test-dev} 2017. The best result
				is highlighted in bold, and same for other tables.}
		\label{COCO val}
	\begin{tabular}{l|c|c|ccccc}
		\toprule
		Method & Back &  size & AP & AP$_{50}$ & AP$_{75}$ & AP$_{M}$ & AP$_{L}$  \\
		\midrule
		\multicolumn{8}{c}{\textbf{Bottom-up methods}}\\
		\hline
		G-RMI\cite{papandreou2017towards}& R101 &353$\times$257& 64.9& 85.5 & 71.3& 62.3 & 70.0\\
		AE\cite{newell2017associative}& - &512$\times$512 &65.5&86.8&72.3&60.6& 72.6\\
		PifPaf\cite{kreiss2019pifpaf} &- &-&67.4& -& - & -&  -\\
		HigherNet\cite{cheng2020higherhrnet:} &HR32&512&66.4& 87.5& 72.8 & 61.2& 74.2 \\
		HGG\cite{jin2020differentiable} &- &512&68.3& 86.7& 75.8 & -& - \\
		FCPose\cite{mao2021fcpose}&R101 &800&65.6& 87.9& 72.6 & 62.1& 72.3 \\
		DEKR\cite{geng2021bottom}&HR32 &512&70.7& 87.7& 77.1 & 66.2& 77.8 \\
		\hline
		\multicolumn{8}{c}{\textbf{Top-down methods}}\\
		\hline
		CPN\cite{chen2018cascaded}& Incep &384$\times$288 &73.0& 91.7&80.9&69.5& 78.1\\
		SBN\cite{xiao2018simple} &R152 &384$\times$288& 73.7& 91.9& 81.1& 70.3& 80.0 \\
		HRNet\cite{sun2019deep} & HR32   &384$\times$288& 74.9&92.5& 82.8& 71.3& 80.9 \\
		PoseFix\cite{Moon2019PoseFixMG}&R152 &384$\times$288& \textbf{76.7}&\textbf{92.6}&\textbf{84.1} &\textbf{73.1} &\textbf{82.6} \\
		UDP\cite{Huang2020TheDI}&R152 &384$\times$288& 74.7&91.8& 82.1 &71.5 &80.8 \\
		\hline
		\multicolumn{8}{c}{\textbf{Transformer-based methods}}\\
		\hline
		PRTR\cite{li2021pose}& HR32   &384$\times$288& 71.7&90.6& 79.6& 67.6& 78.4 \\
		TFPose\cite{mao2021tfpose}& R50   &384$\times$288& 72.2&90.9& 80.1& 69.1& 78.8 \\
		TokenP\cite{li2021tokenpose}& HR32   &256$\times$192& 74.7&89.8& 81.4& 71.3& 81.4 \\
		TransP\cite{yang2021transpose}& HR32   &256$\times$192&73.4 &91.6&81.1 &70.1 &79.3  \\
		\hline
		\multicolumn{8}{c}{\textbf{Only w/ Category Labels}} \\
		\hline
		Baseline & R50&256$\times$192& 34.0  & 41.4  & 36.2   & 31.6 & 36.2   \\
		Baseline & R50&384$\times$288& 35.1  & 42.1  & 37.2  & 32.6 &  37.3 \\
		\hline
		\textbf{Ours-LOFR} & R50 &256$\times$192& 54.3  & 61.0 &  55.6 & 52.6 & 56.7   \\
		\textbf{Ours-LOFR} & R50 &384$\times$288& 55.4  & 62.1 &  56.7 & 53.7  & 57.9   \\
		\hline
		\textbf{Ours-LOFR} & HR32 &256$\times$192& 54.8 &61.8  & 56.1  & 53.2 & 57.3   \\
		\textbf{Ours-LOFR} & HR32 &384$\times$288& 55.9 & 62.9 & 55.3  & 54.4 & 58.4   \\
		\textbf{Ours-LOFR} & HR48 &256$\times$192& 55.5 & 62.4 & 56.7  & 53.8 & 57.8   \\
		\rowcolor{gray!20}
		\textbf{Ours-LOFR} & HR48 &384$\times$288& 56.4 & 63.4 & 55.6  &54.8  & 59.0   \\
		\bottomrule
	\end{tabular}
\end{table}

\noindent\textbf{On MPII.} As shown in Tab.~\ref{tab:2 MPII}, only with the category labels, the baseline solely achieves $35.3\%$ PCKh score. By comparison, the accuracy of ours boosts to $61.8\%$ with a large margin of $26.5\%$.
Additionally, our LOFR achieves consistent improvements on all kinds of joints, though a certain gap exists compared with the fully-supervised methods. It's possibly in that MPII contains in-the-wild images with diverse pose interactions which span from householding to outdoor sports thus it brings great challenges to the model only with category-level labels.

\begin{table}[ht]
	\centering
	\setlength{\tabcolsep}{0.8mm}
	\setlength{\abovecaptionskip}{0cm}
	\setlength{\belowcaptionskip}{0cm}
		\caption{Performance comparisons of PCKh@0.5 on MPII \emph{test} set. * means extra labels and larger image size are used.}
		\label{tab:2 MPII}
	\begin{tabular}{l|ccccccc|c}
		\toprule
		Method & Hea & Sho & Elb & Wri & Hip & Kne & Ank & Total \\
		\midrule
		Wei\cite{wei2016convolutional} & 97.8 & 95.0 & 88.7 & 84.0 & 88.4 & 82.8 & 79.4 & 88.5 \\
		Newell\cite{newell2016stacked} & 98.2 & 96.3& 91.2 & 87.2 & 89.8 & 87.4 & 83.6 & 90.9 \\
		Sun\cite{sun2017human} & 98.1 & 96.2 & 91.2& 87.2 & 89.8 & 87.4 & 84.1 & 91.0 \\
		Tang\cite{tang2018quantized} & 97.4 & 96.4 & 92.1 & 87.7 & 90.2 & 87.7 & 84.3 & 91.2 \\
		Ning\cite{ning2018knowledge-guided} & 98.1 & 96.3 & 92.2 & 87.8 & 90.6 & 87.6 & 82.7 & 91.2 \\
		Chu\cite{chu2017multi-context} & 98.5& 96.3 & 91.9 & 88.1 & 90.6 & 88.0 & 85.0 & 91.5 \\
		Chou\cite{chou2018self} & 98.2& 96.8 & 92.2 & 88.0 & 91.3 & 89.1 & 84.9 & 91.8 \\
		Yang\cite{yang2017learning} & 98.5 & 96.7 & 92.5 & 88.7 & 91.1 & 88.6 & 86.0 & 92.0 \\
		Ke\cite{ke2018multi-scale} & 98.5 & 96.8 & 92.7 & 88.4 & 90.6 & 89.3 & 86.3 & 92.1 \\
		Xiao\cite{xiao2018simple} & 98.5 & 96.6 & 91.9 & 87.6 & 91.1 & 88.1 & 84.1 & 91.5 \\
		Tang\cite{tang2018deeply} & 98.4 & 96.9 & 92.6 & 88.7 & 91.8 & 89.4 & 86.2 & 92.3 \\
		Sun\cite{sun2019deep} & 98.6 & 96.9 & 92.8 & 89.0 & 91.5 &89.0 & 85.7 & 92.3 \\
		\hline
		Su*\cite{su2019cascade}& 98.7 & 97.5 & 94.3 & 90.7 & 93.4 &92.2 & 88.4 & 93.9 \\
		Bin*\cite{bin2020adversarial}& \textbf{98.9} & \textbf{97.6} & \textbf{94.6} & \textbf{91.2} & 93.1 &\textbf{92.7} & 89.1 & \textbf{94.1} \\
	    Bulat*\cite{bulat2020toward}& 98.8 & 97.5 & 94.4 & \textbf{91.2} & \textbf{93.2} & 92.2 & \textbf{89.3} & \textbf{94.1} \\
		\hline
		\multicolumn{9}{c}{\textbf{Only W/ Category Labels}}\\	
		\hline
		Baseline & 53.5 & 73.0 &33.8 & 26.5 & 15.7 & 14.6 & 7.1 & 35.3 \\  
		\rowcolor{gray!20}
		\textbf{Ours-LOFR}& 86.9 & 79.8 & 65.6 &54.1 & 47.7 & 46.2 & 32.1  &  61.8 \\ 
		\bottomrule
	\end{tabular}

\end{table} 

\begin{figure}[htp]
	\centering
	\includegraphics[width=3.2in,height=1.35in]{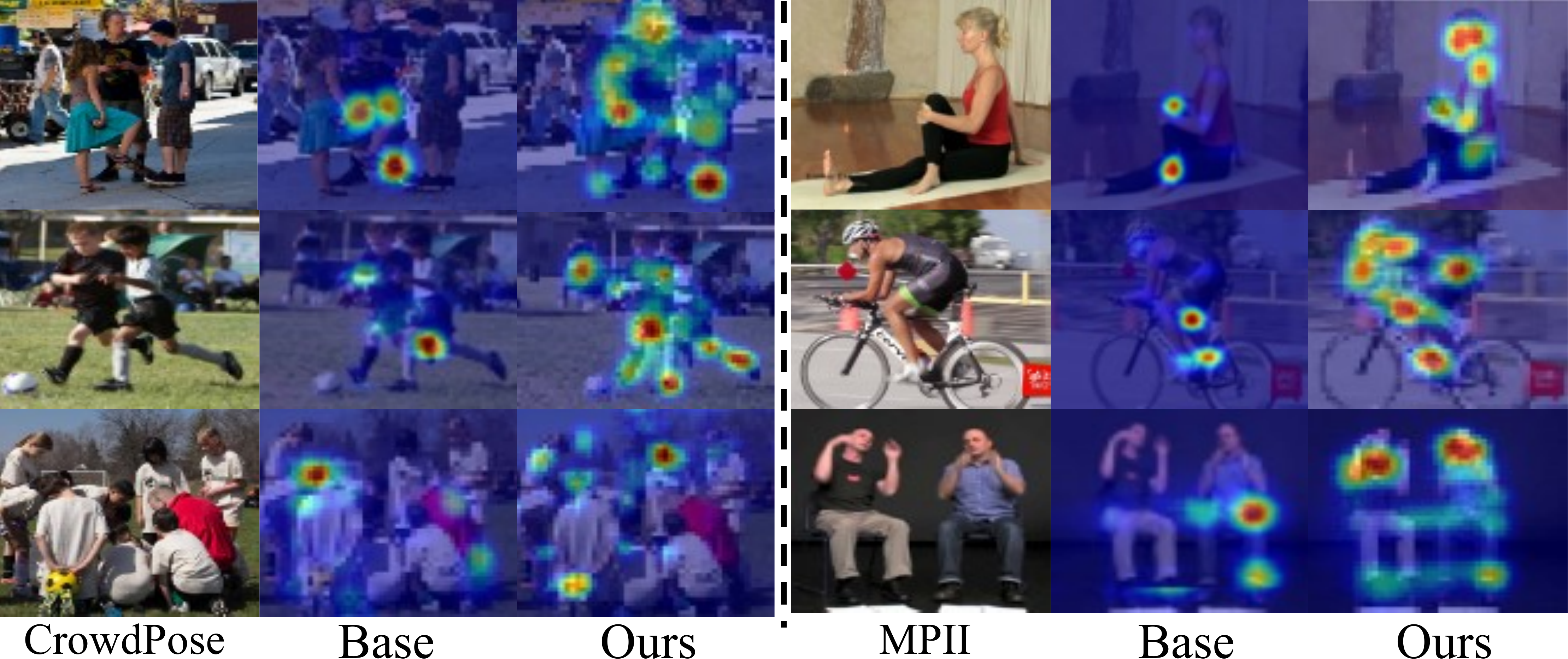}
	\caption{Qualitative comparisons on MPII and CrowdPose.}
	\label{fig:mp_crowd}
\end{figure}

\noindent\textbf{On CrowdPose.} We further validate our method on the challenging CrowdPose dataset and the result is depicted in Tab.~\ref{CrowdPose}. The LOFR surpasses the baseline on all metrics, yielding the accuracy of mAP $42.5\%$ with an improvement of $20.0\%$. Even for the AP \emph{(hard)}, we still bring a large improvement to $34.1\% (+17.5\%)$. This suggests our method is solid even toward the extreme crowded poses. The qualitative comparison results are shown in Fig.~\ref{fig:mp_crowd}.

\begin{table}[ht]
	\centering
	\setlength{\tabcolsep}{1.8mm}
	\setlength{\abovecaptionskip}{0cm}
	\setlength{\belowcaptionskip}{-0.1cm}
		\caption{Performance comparisons on CrowdPose \emph{test} set.}
		\label{CrowdPose}
	\begin{tabular}{l|ccccc}
		\toprule
		Method  &  AP & AP$_{50}$  &AP$_{75}$  & AP$_{M}$ & AP$_{H}$  \\
		\midrule
		\multicolumn{6}{c}{\textbf{Bottom-up methods}}\\
		\hline
		OpenPose\cite{cao2019openpose:} &-  & - & - & 48.7 & 32.3  \\
		HigherHRNet\cite{cheng2020higherhrnet:}  & 67.6  & 87.4  & 72.6  & 68.1   & 58.9\\
		DEKR\cite{geng2021bottom} &  68.0 &85.5 &73.4  & 68.8   &58.4 \\
		\hline
		\multicolumn{6}{c}{\textbf{Top-down methods}}\\
		\hline
		Mask-RCNN\cite{he2017mask}& 57.2& 83.5& 60.3 & 57.9&45.8\\
		SBN\cite{xiao2018simple} & 60.8& 84.2&71.5& 61.2&  51.2 \\
		AlphaPose\cite{li2019crowdpose}&66.0&84.2&71.5& 66.3&57.4\\
		HRNet\cite{sun2019deep} & \textbf{71.7}&\textbf{89.8}&\textbf{76.9} & \textbf{72.7}&\textbf{61.5}\\
		\hline
		Baseline &  21.5 & 39.1&28.2& 23.4 & 16.6\\
		\rowcolor{gray!20}
		\textbf{Ours-LOFR} & 42.5  & 58.7&48.6 & 43.8 & 34.1\\
		\bottomrule
	\end{tabular}

\end{table}

\subsubsection{Weakly Semi-supervised Setting}
While we mainly focus on the image-level learning, our method can also achieve better performance when few location-labeled data are given. In this time, the training process is not changed but the labeled samples conduct supervised loss with groundtruth. We choose ${5\%, 10\%, 25\%}$ labeled instances from the training set of MS-COCO and MPII, the remaining samples only have image-level labels and train as the location-free setting. 

We validate the experiments on the \emph{val} of the above datasets in Tab.~\ref{semi}. “Sup-only” means trained only with the samples labeled with groundtruth. And we reimplement the Sup-only baseline and the compared methods combined with the transformer baseline framework for the fair comparison. It shows that our model achieves stable improvements than the base \emph{Sup-only} in all cases with the category label assistance. Noted that when $25\%$ location labeled instances are given, our model achieves comparable performance with the fully-supervised (\emph{ALL}) model. Moreover, the proposed strategy can benefit more accurate estimation under the fully-supervised setting and it achieves the result of $71.8\%$ and $89.3\%$ on COCO and MPII, outperforming the base (\emph{ALL}) model by $0.8\%$, $0.6\%$, respectively. 

\begin{table}[h]
	\centering
	\setlength{\tabcolsep}{0.39mm}
	\setlength{\abovecaptionskip}{0cm}
	\setlength{\belowcaptionskip}{-0.1cm}
		\caption{Result comparisons on \emph{validation} set of the COCO and MPII datasets for different ratios of location data.}
		\label{semi}
	\begin{tabular}{c|l|c|cccc}
		\toprule
		Dataset& Method  & Back & 5\% & 10\% & 25\%   & ALL   \\
		\midrule
		\multirow{5}{*}{COCO}&Sup-only\cite{xiao2018simple} &R50& 50.3 & 54.8  & 60.8 & 71.0  \\
		&Sup-only\cite{sun2019deep} &HR32&53.8 & 58.9 & 64.6 &  74.9  \\ \cline{2-7}
		&SemiPose\cite{xie2021empirical}&R50& 57.7&61.6& 66.4 & -\\ \cline{2-7}
     	&\textbf{Ours-WS} &R50&60.9&  64.8& \textcolor[rgb]{1.00,0.00,0.00}{70.6} & \textbf{71.8}(\textbf{+0.8\%})\\
		&\textbf{Ours-WS} & HR32 &64.6& 68.9&\textcolor[rgb]{1.00,0.00,0.00}{74.0} & \textbf{75.3}(\textbf{+0.5\%})\\
		\hline
		\multirow{3}{*}{MPII}&Sup-only\cite{xiao2018simple} &R50& 64.0 & 69.5  & 77.5 & 88.7  \\ \cline{2-7}
		&SemiPose\cite{xie2021empirical}&R50& 71.3&76.3& 82.5 & -\\ 
		&\textbf{Ours-WS} &R50& 74.6 & 79.8 & \textcolor[rgb]{1.00,0.00,0.00}{88.1} & \textbf{89.3}(\textbf{+0.6\%})\\
		\bottomrule
	\end{tabular}

\end{table}

\subsection{Ablation Study}
\noindent\textbf{Effectiveness of SPE, MS and PDC.} In Tab.~\ref{component}, the model with SPE improves $3.4\%$ mAP than baseline. This reveals that it's useful to let model focus on the local part context learning. And multi-scale (MS) strategy further improves $1.8\%$ and the PDC also improves $2.0\%$ based on RePPG. 

\begin{figure}[h]
	\centering
	\includegraphics[width=3.2in,height=1.2in]{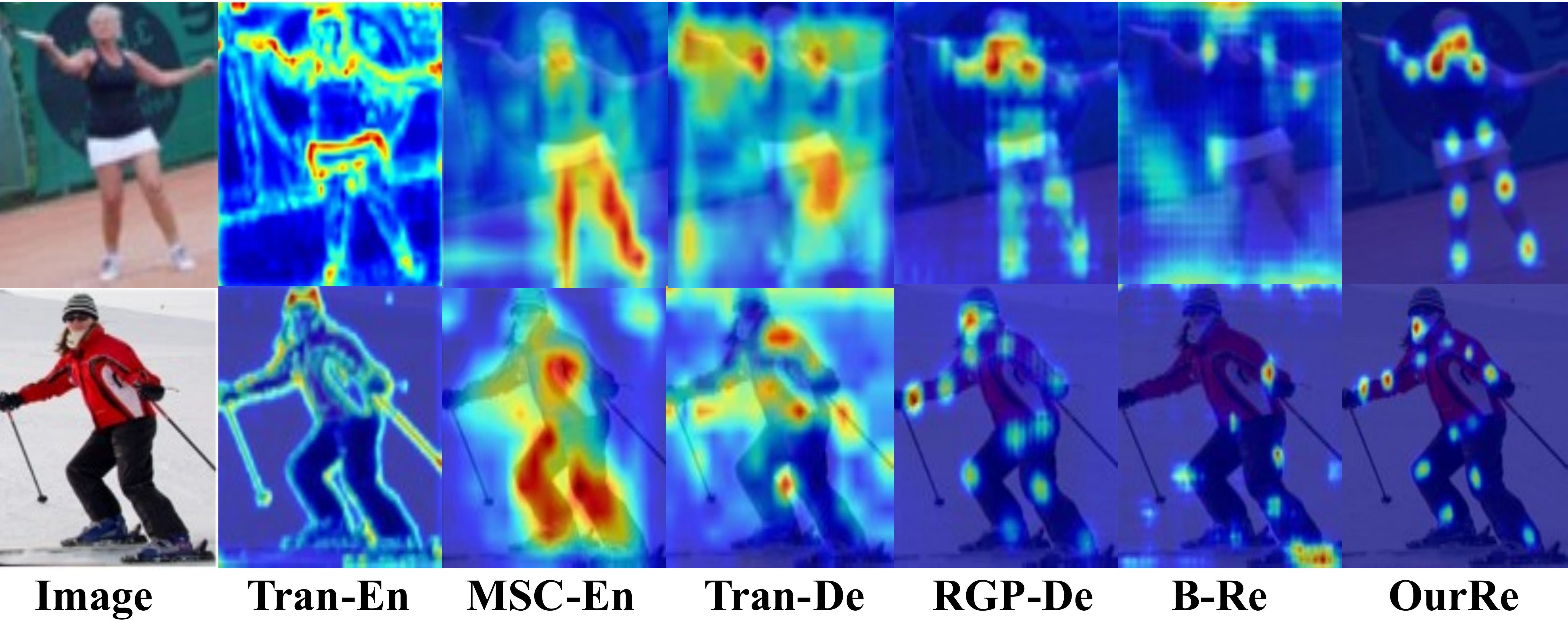}
	\caption{The original transformer-based encoder (Tran-En) vs. the proposed encoder (MSC-En) and the original decoder (Tran-De) vs. the proposed decoder (RGP-De). B-Re and OurRe depict the result of baseline and ours, respectively.}
	\label{fig:abla}
\end{figure}

For further showing its effectiveness intuitively, we visualize the location maps produced by the model with or without SPE and PDC in Fig.~\ref{fig:ablation}. We observe that with SPE, the keypoint location is more accurate and complete, which illustrating our SPE provides spatial-aware guidance compared with the random initialized position encoding. The PDC also helps model discover more explicit part regions than baseline. Additionally, the learned features by MSC-En can cover more part-specific context by the aid of SPE compared with the Tran-En in Fig.~\ref{fig:abla}.

\begin{table}[htbp]
	\centering
	\setlength{\abovecaptionskip}{0cm}
	\setlength{\belowcaptionskip}{-0.1cm}
	\setlength{\tabcolsep}{1.1mm}
			\caption{Ablation study of each module on COCO \emph{val2017}.}
	\label{component}
	\begin{tabular}{c|ccccc|cc}
		\hline
		Model & Baseline & SPE &MS& RePPG & PDC & mAP  & mAR  \\
		\hline 
		1 &$\checkmark$&  &  & &  &39.1  &44.5 \\
		2 &$\checkmark$&  $\checkmark$  & &  &   & 42.5  & 47.8  \\
		3 &$\checkmark$&   &$\checkmark$ &  &   & 40.9  & 46.7  \\
		4 &$\checkmark$ & && $\checkmark$ &  & 44.9  & 50.4   \\
		5 & $\checkmark$& & && $\checkmark$ &  41.1  & 46.5\\
		6 &$\checkmark$& $\checkmark$  &$\checkmark$ &  &   & 44.3  & 49.4  \\
		7 & $\checkmark$& & &$\checkmark$& $\checkmark$ &  46.9  &52.6 \\ \hline
		8 & $\checkmark$&$\checkmark$ &$\checkmark$ & $\checkmark$& $\checkmark$ & $\textbf{54.9}$  & $\textbf{60.6}$\\
		\hline
	\end{tabular}

\end{table}

\begin{figure}[htp]
	\centering
	\setlength{\abovecaptionskip}{0.15cm}
	\setlength{\belowcaptionskip}{0cm}
	\includegraphics[width=3.2in,height=1.6in]{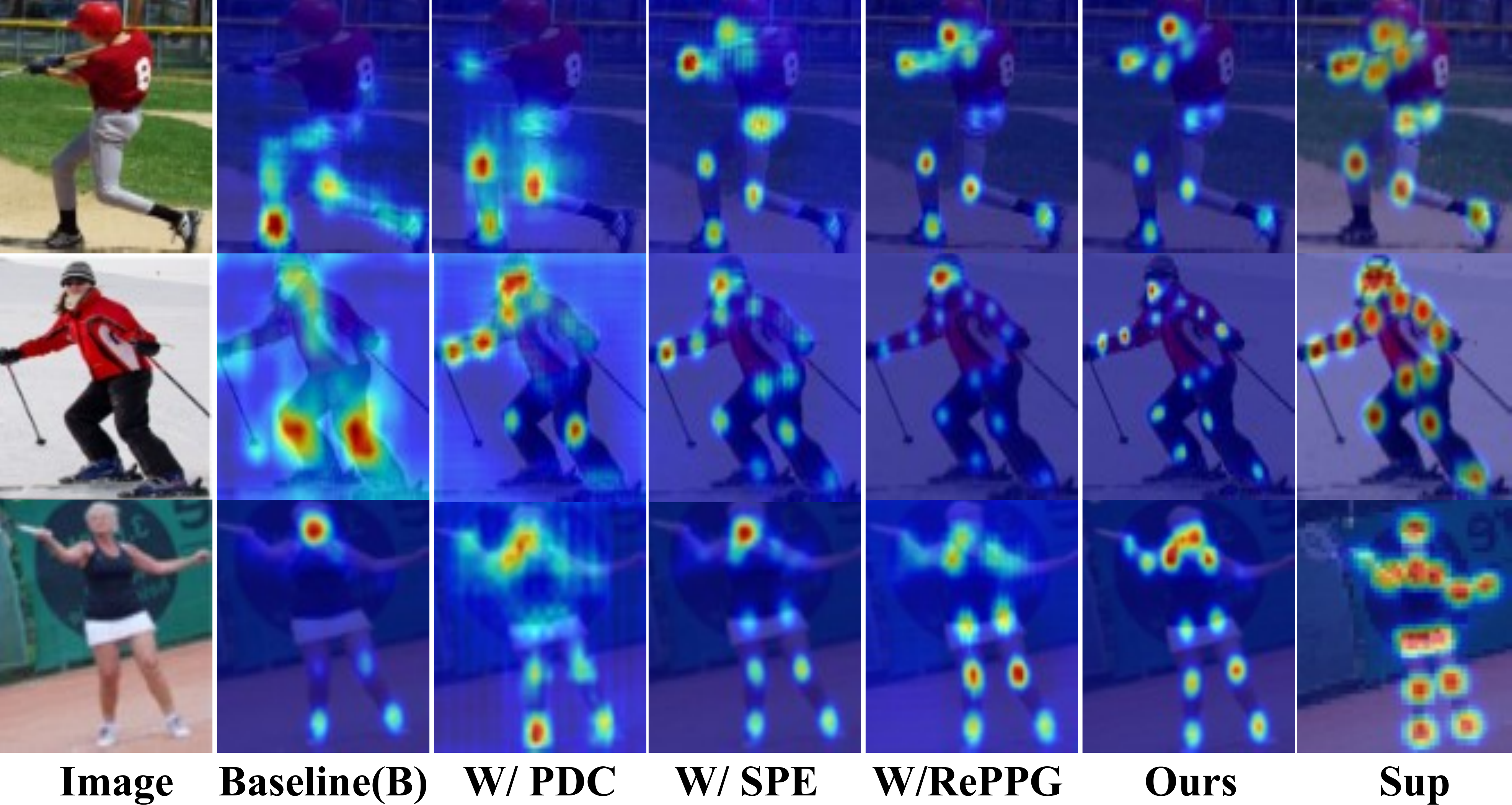}
	\caption{The ablated qualitative results for each module.}
	\label{fig:ablation}
\end{figure}

\noindent\textbf{Effectiveness of RePPG.} In Tab.~\ref{component}, the result of mAP improves to $44.9\% (+5.8\%)$ by adding RePPG on the baseline. This powerfully indicates the necessity of mining the potential structural keypoint relations for guidance. 

\begin{figure}[htp]
	\centering
	\includegraphics[width=3.3in,height=2.6in]{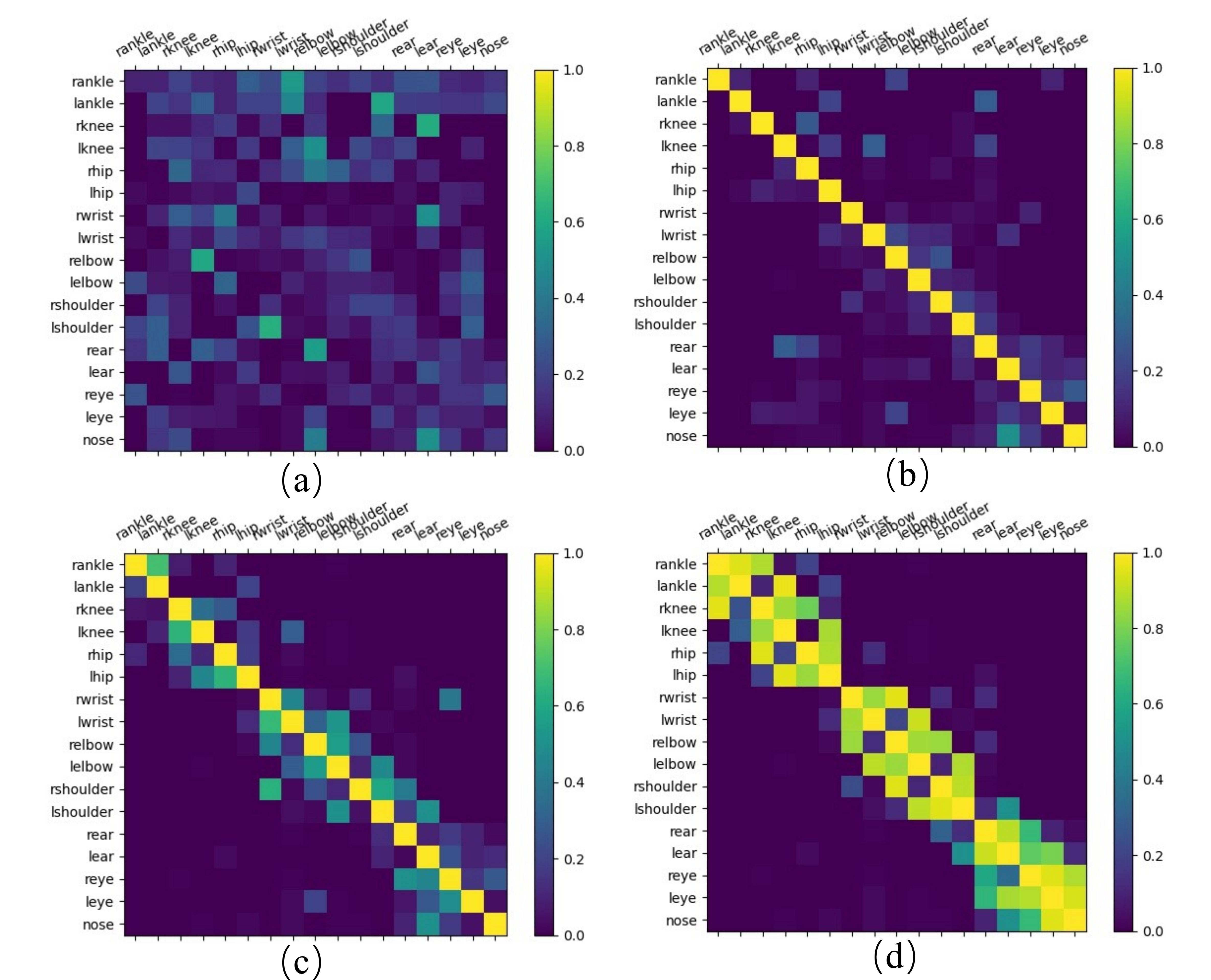}
	\caption{The inner product matrix of the learned pose prototypes. Each row of the matrix denotes the learned prior relations for a given type of keypoint with others. (a) depicts the random initialization; (b) depicts processed by SA; (c) the initial joint relevance; (d) depicts learned by RePPG.}
	\label{fig:gcn}
\end{figure}

In our setting, the pose prototypes learn the statistical relevance between keypoints from the dataset, serving as the prior knowledge. To indicate the information encoded in these prototypes, we calculate their inner product matrix and visualize it in different cases in Fig.~\ref{fig:gcn}. Fig.~\ref{fig:gcn} (d) reveals that one tends to be highly related to its symmetric or adjacent keypoints. For instance, the left hip is mostly related to right hip and left shoulder with high relevance score. Such finding conforms to our common sense and reveals what the model learns. But the Fig.~\ref{fig:gcn} (b) mainly implies the self-correlation learning, which indicates that RePPG can encode more explicit local keypoint relations.

\noindent\textbf{Impact of the model scaling.} In Tab.~\ref{layer}, we explore the effect of numbers of encoders and decoders in transformer. The performance grows at the first four layers and saturates as the layer increases and we choose the best setting.
\begin{table}[ht]
	\centering
	\setlength{\abovecaptionskip}{0cm}
	\setlength{\belowcaptionskip}{0cm}
	\caption{Effect of encoder ($D_{En}$) and decoder ($D_{De}$) numbers on COCO \emph{val2017}.}
	\label{layer}
	\begin{tabular}{c|cc|cc}
		\hline
		Model &$D_{En}$ & $D_{De}$ & mAP & mAR  \\
		\hline
		1 & 2 &2  &50.3  & 55.2 \\
		2 & 3 &3  &52.3  & 57.8 \\
		3 & 4 &4  &53.7  & 58.7 \\
		4& 4  & 6& \textbf{54.9} & \textbf{60.6}  \\
		5& 6  & 4& 54.4 & 59.6  \\
		6 & 6 & 6  & 53.2 & 58.4 \\  
		\hline
	\end{tabular}

\end{table}
 
\begin{figure}[h]
	\centering
	\includegraphics[width=3.2in,height=1.6in]{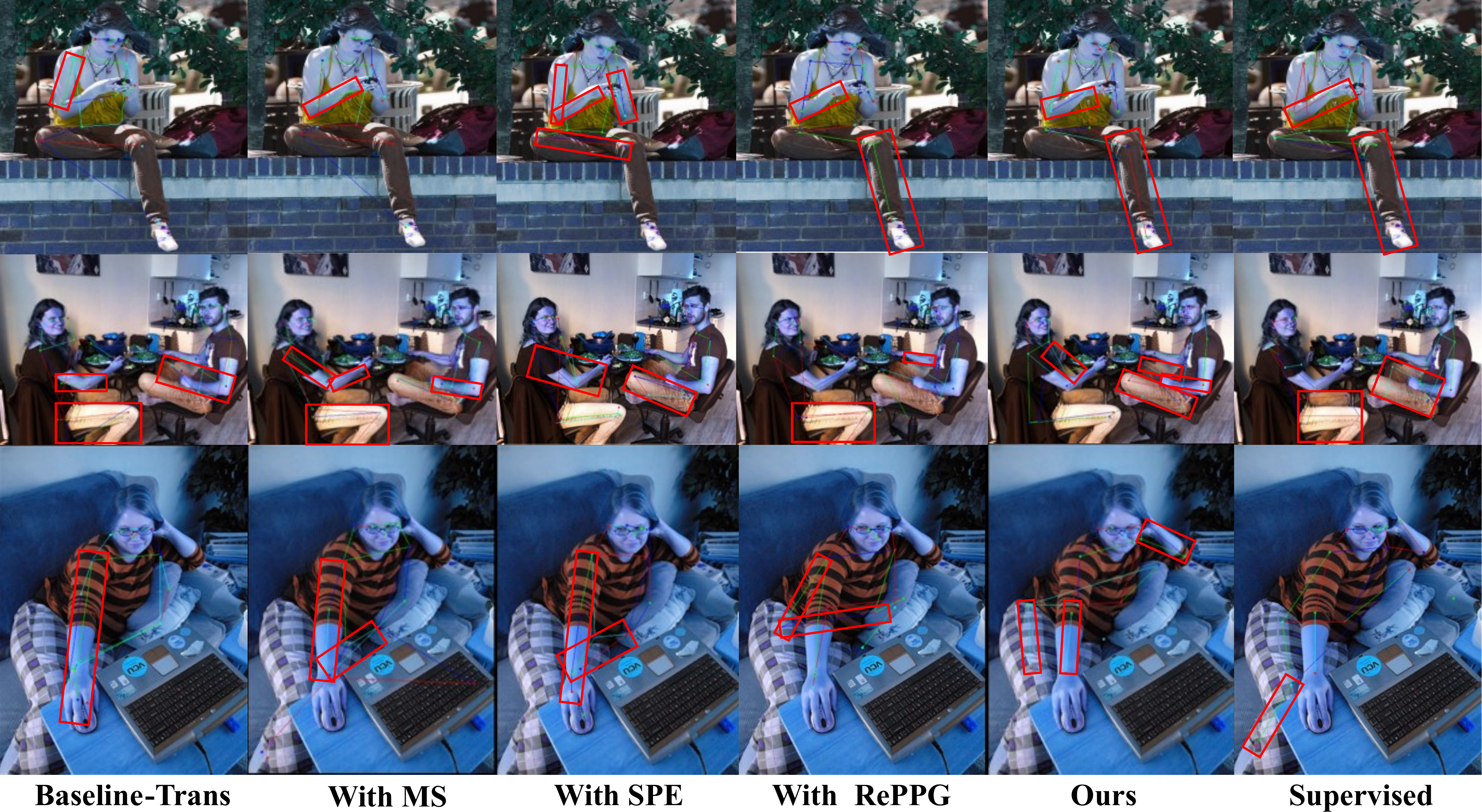}
	\caption{The qualitative results on random selected samples. We highlight the notable differences by a red rectangle box.}
	\label{fig:comparison}
\end{figure}

\noindent\textbf{Visualization and Analysis.} For illustrating our strategy intuitively, we visualize the detailed process in heatmap-based results in Fig.~\ref{fig:abla}. Our RGP-De decodes more fine-grained and distinguished keypoint feature responses than the baseline. Our final result acquires more accurate locations even compared with the supervised result as in Fig.~\ref{fig:ablation}. Besides, we also visualize the qualitative comparison results in Fig.~\ref{fig:comparison} and it also can prove the effectiveness of our method for the weakly-supervised HPE. 

For ensuring the results are not cherry-picked, we compute the statistical average response value across joints on the whole dataset in Tab.~\ref{tab:response}. The joint response value of LOFR is obviously higher than the baseline. This illustrates that our model can acquire more explicit joint locations by the aid of the learned local context and guided relations. 

\begin{table}[ht]
	\centering
	\setlength{\abovecaptionskip}{0.1cm}
	\setlength{\belowcaptionskip}{0cm}
	\setlength{\tabcolsep}{0.47mm}
	\caption{Comparisons of the average response around the body joints across datasets. We normalize the value belongs to $[0\sim1]$.}
	\label{tab:response}
	\begin{tabular}{c|cccccccccc}
		\toprule
		Dataset & Ear& Eye&Nose & Hea & Sho & Elb & Wri & Hip & Kne & Ank \\
		\midrule
		\multicolumn{10}{c}{\textbf{Baseline}}\\	
		\hline
		COCO & 0.21  & 0.17 &0.32  & 0.65 &0.36 & 0.31 &0.21 &0.39 &0.21 & 0.13  \\
		\hline
		MPII & - & - & - & 0.62 & 0.52 & 0.34 & 0.24 & 0.17& 0.16& 0.09  \\
		\hline
		Crowd & - & - & -& 0.61 & 0.32 & 0.28 & 0.16 & 0.35 & 0.15 &0.11   \\
		\hline
		\multicolumn{10}{c}{\textbf{Ours-LOFR}}\\
		\hline
		COCO &0.45  & 0.38& 0.52 & 0.86 & 0.71 & 0.53 &0.46 &0.61 &0.54 & 0.42  \\
		\hline
		MPII  & - & - & - & 0.85 & 0.76 & 0.64 &0.53 &0.48 &0.50 &0.35   \\
		\hline
		Crowd & - & - & -& 0.82 & 0.62 &0.51 &0.42& 0.51& 0.42& 0.35  \\ 
		\bottomrule
	\end{tabular}
\end{table} 
 
\noindent\textbf{Discussions.} 
Benefiting from the global feature learning capability of transformer, we can obtain more discriminative and comprehensive human body context than CNN. However, the body parts are very small and hard to distinguish, it's vital to enable the transformer possess the position-aware prior to know where should be attended. Thus, we design the spatial-aware position encoding to focus locally. 
More significantly, the structural relations encoded via GCN can guide the decoder to activate the delicate part locations. With their collaboration, we achieve a relatively competitive location-free HPE.

Considering the issues of the image-label similarity, when a person appears completely, it possesses all keypoint categories. However, for the dataset we adopt, \emph{e.g.}, COCO, according to the statistics in~\cite{ruggero2017benchmarking}, the complete instances account for less than $50\%$, most instances only have half or a small number of body parts appearing in images. Therefore, the category labels across instances have sufficient diversity to enable model capture the distinct joint information.

\section{Conclusion and Limitation}
In this paper, we shift the paradigm of human pose estimation from \emph{Location-supervised} to \emph{Location-free}. Accordingly, we propose a customized transformer-based HPE pipeline from the perspective of classification only with category-level labels.We firstly design a multi-scale spatial-guided context encoder to capture comprehensive context while focusing on the local part regions. To decode more accurate part-aware locations, we consider the inherent relation constraint among joints via GCN to encode the relational guidance into the pose prototypes. A part diversity constraint is used to keep the part features distinguished.  

\noindent\textbf{Limitation.} The complexity of the model can be further reduced. And the model cannot better solve the occlusions for the multi-person estimation especially in crowded scenes (\emph{e.g.}, on CrowdPose) and it worth exploring in future.

{\small
	\bibliographystyle{ieee_fullname}
	\bibliography{egbib}

\begin{thebibliography}{10}\itemsep=-1pt

\bibitem{andriluka20142d}
Mykhaylo Andriluka, Leonid Pishchulin, Peter~V Gehler, and Bernt Schiele.
\newblock 2d human pose estimation: New benchmark and state of the art
  analysis.
\newblock {\em CVPR}, pages 3686--3693, 2014.

\bibitem{bin2020adversarial}
Yanrui Bin, Xuan Cao, Xinya Chen, Yanhao Ge, Ying Tai, Chengjie Wang, Jilin Li,
  Feiyue Huang, Changxin Gao, and Nong Sang.
\newblock Adversarial semantic data augmentation for human pose estimation.
\newblock {\em European Conference on Computer Vision}, pages 606--622, 2020.

\bibitem{bulat2020toward}
Adrian Bulat, Jean Kossaifi, Georgios Tzimiropoulos, and Maja Pantic.
\newblock Toward fast and accurate human pose estimation via soft-gated skip
  connections.
\newblock {\em 2020 15th IEEE International Conference on Automatic Face and
  Gesture Recognition (FG 2020)}, pages 8--15, 2020.

\bibitem{cao2019openpose:}
Zhe Cao, Gines Hidalgo, Tomas Simon, Shihen Wei, and Yaser Sheikh.
\newblock Openpose: Realtime multi-person 2d pose estimation using part
  affinity fields.
\newblock {\em IEEE Transactions on Pattern Analysis and Machine Intelligence},
  pages 1--1, 2019.

\bibitem{carion2020end}
Nicolas Carion, Francisco Massa, Gabriel Synnaeve, Nicolas Usunier, Alexander
  Kirillov, and Sergey Zagoruyko.
\newblock End-to-end object detection with transformers.
\newblock In {\em ECCV}, pages 213--229. Springer, 2020.

\bibitem{chen2018cascaded}
Yilun Chen, Zhicheng Wang, Yuxiang Peng, Zhiqiang Zhang, Gang Yu, and Jian Sun.
\newblock Cascaded pyramid network for multi-person pose estimation.
\newblock {\em CVPR}, pages 7103--7112, 2018.

\bibitem{cheng2020higherhrnet:}
Bowen Cheng, Bin Xiao, Jingdong Wang, Honghui Shi, Thomas~S Huang, and Lei
  Zhang.
\newblock Higherhrnet: Scale-aware representation learning for bottom-up human
  pose estimation.
\newblock In {\em Proceedings of the IEEE/CVF conference on computer vision and
  pattern recognition}, pages 5386--5395, 2020.

\bibitem{chou2018self}
Chiajung Chou, Juiting Chien, and Hwanntzong Chen.
\newblock Self adversarial training for human pose estimation.
\newblock {\em Asia Pacific Signal and Information Processing Association
  Annual Summit and Conference}, pages 17--30, 2018.

\bibitem{chu2017multi-context}
Xiao Chu, Wei Yang, Wanli Ouyang, Cheng Ma, Alan~L Yuille, and Xiaogang Wang.
\newblock Multi-context attention for human pose estimation.
\newblock {\em CVPR}, pages 5669--5678, 2017.

\bibitem{dosovitskiy2020image}
Alexey Dosovitskiy, Lucas Beyer, Alexander Kolesnikov, Dirk Weissenborn,
  Xiaohua Zhai, Thomas Unterthiner, Mostafa Dehghani, Matthias Minderer, Georg
  Heigold, Sylvain Gelly, et~al.
\newblock An image is worth 16x16 words: Transformers for image recognition at
  scale.
\newblock {\em ICLR}, 2020.

\bibitem{geng2021bottom}
Zigang Geng, Ke Sun, Bin Xiao, Zhaoxiang Zhang, and Jingdong Wang.
\newblock Bottom-up human pose estimation via disentangled keypoint regression.
\newblock {\em Proceedings of the IEEE/CVF Conference on Computer Vision and
  Pattern Recognition}, pages 14676--14686, 2021.

\bibitem{he2017mask}
Kaiming He, Georgia Gkioxari, Piotr Dollar, and Ross Girshick.
\newblock Mask r-cnn.
\newblock {\em ICCV}, pages 2980--2988, 2017.

\bibitem{Huang2020TheDI}
Junjie Huang, Z. Zhu, F. Guo, and G. Huang.
\newblock The devil is in the details: Delving into unbiased data processing
  for human pose estimation.
\newblock {\em 2020 IEEE/CVF Conference on Computer Vision and Pattern
  Recognition (CVPR)}, pages 5699--5708, 2020.

\bibitem{huang2020hot}
Lin Huang, Jianchao Tan, Jingjing Meng, Ji Liu, and Junsong Yuan.
\newblock Hot-net: Non-autoregressive transformer for 3d hand-object pose
  estimation.
\newblock In {\em ACM Multimedia}, pages 3136--3145, 2020.

\bibitem{jin2020differentiable}
Sheng Jin, Wentao Liu, Enze Xie, Wenhai Wang, Chen Qian, Wanli Ouyang, and Ping
  Luo.
\newblock Differentiable hierarchical graph grouping for multi-person pose
  estimation.
\newblock {\em European Conference on Computer Vision}, pages 718--734, 2020.

\bibitem{ke2018multi-scale}
Lipeng Ke, Mingching Chang, Honggang Qi, and Siwei Lyu.
\newblock Multi-scale structure-aware network for human pose estimation.
\newblock {\em ECCV}, pages 731--746, 2018.

\bibitem{kingma2015adam:}
Diederik~P Kingma and Jimmy Ba.
\newblock Adam: A method for stochastic optimization.
\newblock {\em ICLR}, 2015.

\bibitem{kocabas2018multiposenet:}
Muhammed Kocabas and Emre Akbas.
\newblock Multiposenet: Fast multi-person pose estimation using pose residual
  network.
\newblock {\em ECCV}, pages 437--453, 2018.

\bibitem{kreiss2019pifpaf}
Sven Kreiss, Lorenzo Bertoni, and Alexandre Alahi.
\newblock Pifpaf: Composite fields for human pose estimation.
\newblock {\em Proceedings of the IEEE/CVF Conference on Computer Vision and
  Pattern Recognition}, pages 11977--11986, 2019.

\bibitem{kumar2017hide}
Krishna Kumar~Singh and Yong Jae~Lee.
\newblock Hide-and-seek: Forcing a network to be meticulous for
  weakly-supervised object and action localization.
\newblock In {\em ICCV}, pages 3524--3533, 2017.

\bibitem{li2019crowdpose}
Jiefeng Li, Can Wang, Hao Zhu, Yihuan Mao, Hao-Shu Fang, and Cewu Lu.
\newblock Crowdpose: Efficient crowded scenes pose estimation and a new
  benchmark.
\newblock In {\em Proceedings of the IEEE/CVF conference on computer vision and
  pattern recognition}, pages 10863--10872, 2019.

\bibitem{li2021pose}
Ke Li, Shijie Wang, Xiang Zhang, Yifan Xu, Weijian Xu, and Zhuowen Tu.
\newblock Pose recognition with cascade transformers.
\newblock In {\em CVPR}, pages 1944--1953, 2021.

\bibitem{li2021lifting}
Wenhao Li, Hong Liu, Runwei Ding, Mengyuan Liu, and Pichao Wang.
\newblock Lifting transformer for 3d human pose estimation in video.
\newblock {\em arXiv preprint arXiv:2103.14304}, 2021.

\bibitem{li2021tokenpose}
Yanjie Li, Shoukui Zhang, Zhicheng Wang, Sen Yang, Wankou Yang, Shu-Tao Xia,
  and Erjin Zhou.
\newblock Tokenpose: Learning keypoint tokens for human pose estimation.
\newblock In {\em Proceedings of the IEEE/CVF International Conference on
  Computer Vision}, pages 11313--11322, 2021.

\bibitem{lin2014microsoft}
Tsungyi Lin, Michael Maire, Serge~J Belongie, James Hays, Pietro Perona, Deva
  Ramanan, Piotr Dollar, and C~Lawrence Zitnick.
\newblock Microsoft coco: Common objects in context.
\newblock {\em ECCV}, pages 740--755, 2014.

\bibitem{mao2021tfpose}
Weian Mao, Yongtao Ge, Chunhua Shen, Zhi Tian, Xinlong Wang, and Zhibin Wang.
\newblock Tfpose: Direct human pose estimation with transformers.
\newblock {\em arXiv preprint arXiv:2103.15320}, 2021.

\bibitem{mao2021fcpose}
Weian Mao, Zhi Tian, Xinlong Wang, and Chunhua Shen.
\newblock Fcpose: Fully convolutional multi-person pose estimation with dynamic
  instance-aware convolutions.
\newblock In {\em Proceedings of the IEEE/CVF Conference on Computer Vision and
  Pattern Recognition}, pages 9034--9043, 2021.

\bibitem{Moon2019PoseFixMG}
Gyeongsik Moon, Ju~Yong Chang, and Kyoung~Mu Lee.
\newblock Posefix: Model-agnostic general human pose refinement network.
\newblock {\em 2019 IEEE/CVF Conference on Computer Vision and Pattern
  Recognition (CVPR)}, pages 7765--7773, 2019.

\bibitem{newell2016stacked}
Alejandro Newell and Jia Deng.
\newblock Stacked hourglass networks for human pose estimation.
\newblock {\em ECCV}, pages 483--499, 2016.

\bibitem{newell2017associative}
Alejandro Newell, Zhiao Huang, and Jia Deng.
\newblock Associative embedding: End-to-end learning for joint detection and
  grouping.
\newblock {\em NIPS}, pages 2277--2287, 2017.

\bibitem{ning2018knowledge-guided}
Guanghan Ning, Zhi Zhang, and Zhihai He.
\newblock Knowledge-guided deep fractal neural networks for human pose
  estimation.
\newblock {\em IEEE Transactions on Multimedia}, pages 1246--1259, 2018.

\bibitem{papandreou2017towards}
George Papandreou, Tyler Zhu, Nori Kanazawa, Alexander Toshev, Jonathan
  Tompson, Chris Bregler, and Kevin~P Murphy.
\newblock Towards accurate multi-person pose estimation in the wild.
\newblock {\em CVPR}, pages 3711--3719, 2017.

\bibitem{Paszke2017AutomaticDI}
Adam Paszke, Sam Gross, Soumith Chintala, Gregory Chanan, Edward Yang, Zachary
  DeVito, Zeming Lin, Alban Desmaison, Luca Antiga, and Adam Lerer.
\newblock Automatic differentiation in pytorch.
\newblock {\em NIPS}, 2017.

\bibitem{ruggero2017benchmarking}
Matteo Ruggero~Ronchi and Pietro Perona.
\newblock Benchmarking and error diagnosis in multi-instance pose estimation.
\newblock {\em Proceedings of the IEEE international conference on computer
  vision}, pages 369--378, 2017.

\bibitem{su2019cascade}
Zhihui Su, Ming Ye, Guohui Zhang, Lei Dai, and Jianda Sheng.
\newblock Cascade feature aggregation for human pose estimation.
\newblock {\em arXiv preprint arXiv:1902.07837}, 2019.

\bibitem{sun2017human}
Ke Sun, Cuiling Lan, Junliang Xing, Wenjun Zeng, Dong Liu, and Jingdong Wang.
\newblock Human pose estimation using global and local normalization.
\newblock {\em ICCV}, pages 5600--5608, 2017.

\bibitem{sun2019deep}
Ke Sun, Bin Xiao, Dong Liu, and Jingdong Wang.
\newblock Deep high-resolution representation learning for human pose
  estimation.
\newblock {\em CVPR}, pages 5693--5703, 2019.

\bibitem{tang2018deeply}
Wei Tang, Pei Yu, and Ying Wu.
\newblock Deeply learned compositional models for human pose estimation.
\newblock {\em ECCV}, pages 197--214, 2018.

\bibitem{tang2018quantized}
Zhiqiang Tang, Xi Peng, Shijie Geng, Lingfei Wu, Shaoting Zhang, and Dimitris~N
  Metaxas.
\newblock Quantized densely connected u-nets for efficient landmark
  localization.
\newblock {\em ECCV}, pages 348--364, 2018.

\bibitem{vaswani2017attention}
Ashish Vaswani, Noam Shazeer, Niki Parmar, Jakob Uszkoreit, Llion Jones,
  Aidan~N Gomez, Lukasz Kaiser, and Illia Polosukhin.
\newblock Attention is all you need.
\newblock {\em arXiv preprint arXiv:1706.03762}, 2017.

\bibitem{wei2016convolutional}
Shihen Wei, Varun Ramakrishna, Takeo Kanade, and Yaser Sheikh.
\newblock Convolutional pose machines.
\newblock {\em CVPR}, pages 4724--4732, 2016.

\bibitem{xiao2018simple}
Bin Xiao, Haiping Wu, and Yichen Wei.
\newblock Simple baselines for human pose estimation and tracking.
\newblock {\em ECCV}, pages 472--487, 2018.

\bibitem{xie2021empirical}
Rongchang Xie, Chunyu Wang, Wenjun Zeng, and Yizhou Wang.
\newblock An empirical study of the collapsing problem in semi-supervised 2d
  human pose estimation.
\newblock {\em Proceedings of the IEEE/CVF International Conference on Computer
  Vision}, pages 11240--11249, 2021.

\bibitem{yang2021transpose}
Sen Yang, Zhibin Quan, Mu Nie, and Wankou Yang.
\newblock Transpose: Keypoint localization via transformer.
\newblock {\em Proceedings of the IEEE/CVF International Conference on Computer
  Vision}, pages 11802--11812, 2021.

\bibitem{yang2017learning}
Wei Yang, Shuang Li, Wanli Ouyang, Hongsheng Li, and Xiaogang Wang.
\newblock Learning feature pyramids for human pose estimation.
\newblock {\em ICCV}, pages 1290--1299, 2017.

\bibitem{yun2019cutmix}
Sangdoo Yun, Dongyoon Han, Seong~Joon Oh, Sanghyuk Chun, Junsuk Choe, and
  Youngjoon Yoo.
\newblock Cutmix: Regularization strategy to train strong classifiers with
  localizable features.
\newblock In {\em Proceedings of the IEEE/CVF international conference on
  computer vision}, pages 6023--6032, 2019.

\bibitem{zhang2021complementary}
Fei Zhang, Chaochen Gu, Chenyue Zhang, and Yuchao Dai.
\newblock Complementary patch for weakly supervised semantic segmentation.
\newblock pages 7242--7251, 2021.

\bibitem{zhang2020inter}
Xiaolin Zhang, Yunchao Wei, and Yi Yang.
\newblock Inter-image communication for weakly supervised localization.
\newblock {\em ECCV}, pages 271--287, 2020.

\bibitem{zhao2019semantic}
Long Zhao, Xi Peng, Yu Tian, Mubbasir Kapadia, and Dimitris~N Metaxas.
\newblock Semantic graph convolutional networks for 3d human pose regression.
\newblock {\em CVPR}, pages 3425--3435, 2019.

\bibitem{zheng20213d}
Ce Zheng, Sijie Zhu, Matias Mendieta, Taojiannan Yang, Chen Chen, and Zhengming
  Ding.
\newblock 3d human pose estimation with spatial and temporal transformers.
\newblock {\em arXiv preprint arXiv:2103.10455}, 2021.

\bibitem{zheng2020end}
Minghang Zheng, Peng Gao, Xiaogang Wang, Hongsheng Li, and Hao Dong.
\newblock End-to-end object detection with adaptive clustering transformer.
\newblock {\em arXiv preprint arXiv:2011.09315}, 2020.

\bibitem{zheng2021rethinking}
Sixiao Zheng, Jiachen Lu, Hengshuang Zhao, Xiatian Zhu, Zekun Luo, Yabiao Wang,
  Yanwei Fu, Jianfeng Feng, Tao Xiang, Philip~HS Torr, et~al.
\newblock Rethinking semantic segmentation from a sequence-to-sequence
  perspective with transformers.
\newblock In {\em CVPR}, pages 6881--6890, 2021.

\bibitem{zhou2016learning}
Bolei Zhou, Aditya Khosla, Agata Lapedriza, Aude Oliva, and Antonio Torralba.
\newblock Learning deep features for discriminative localization.
\newblock In {\em CVPR}, pages 2921--2929, 2016.

\bibitem{zhu2020deformable}
Xizhou Zhu, Weijie Su, Lewei Lu, Bin Li, Xiaogang Wang, and Jifeng Dai.
\newblock Deformable detr: Deformable transformers for end-to-end object
  detection.
\newblock {\em ICLR}, 2020.

\end{thebibliography}
}

\end{document}